\documentclass{scrartcl}
\usepackage{ceurart}
\usepackage{listings}
\usepackage{amsmath,amssymb}
\usepackage{booktabs}
\usepackage{xcolor}
\usepackage{multirow}
\usepackage{array}
\usepackage{microtype}
\usepackage{tikz}
\usetikzlibrary{arrows.meta,shapes.geometric,positioning,calc}
\usepackage{algorithm}
\usepackage{algpseudocode}

\newcommand{\diagM}{\ensuremath{\operatorname{diag}(M)}}
\newcommand{\agopw}{AGOP\mbox{-}Weighted}
\newcommand{\agopg}{AGOP\mbox{-}Global}
\newcommand{\agopl}{AGOP\mbox{-}Local}

\begin{document}

\copyrightyear{2026} 
\copyrightclause{Copyright for this paper by its authors.
  Use permitted under Creative Commons License Attribution 4.0
  International (CC BY 4.0).}
\conference{Late-breaking work, Demos and Doctoral Consortium, colocated with the 4th World Conference on eXplainable Artificial Intelligence: July 01--03, 2026, Fortaleza, Brazil}

\title{AGOP as Explanation: From Feature Learning to Per-Sample Attribution in Image Classifiers}

\author[1]{Raj Kiran Gupta Katakam}[%
email=raj.katakam@creditkarma.com,
]
\cormark[1]

\address[1]{Credit Karma, San Francisco, CA, USA}

\cortext[1]{Corresponding author.}

\begin{abstract}
The Average Gradient Outer Product (AGOP) governs feature learning in neural
networks: the Neural Feature Ansatz states that weight Gram matrices at each
layer align with the corresponding AGOP matrices computed over the training
distribution.
We ask a complementary question: can this same quantity serve as a
\emph{post-hoc attribution method} for explaining individual predictions?
We introduce \emph{\agopw{}}: a novel attribution method that multiplies the
per-sample gradient by $\sqrt{\diagM/\max\diagM}$, a training-distribution
prior that suppresses gradient noise and amplifies consistently important
pixels---a combination not present in any prior attribution method.
We formalise two companion variants---\agopl{} (per-sample gradient, equivalent
to VanillaGrad) and \agopg{} ($\diagM$ directly as a zero-cost saliency
map)---and implement an efficient training-time accumulation hook; \agopg{} then requires zero inference cost (disk lookup) while \agopw{} requires only a single gradient pass.
We conduct the first rigorous comparison of AGOP attribution against
Integrated Gradients (IG), SmoothGrad, GradCAM, and VanillaGrad across two
benchmarks with pixel-level ground truth:
(i)~the synthetic XAI-TRIS benchmark (four classification scenarios, 8$\times$8
images, CNN8by8) and
(ii)~the photorealistic CLEVR-XAI benchmark (ResNet-18 fine-tuned from
ImageNet).
\agopw{} achieves 44\% higher mIoU than IG on linear tasks; \agopg{} achieves
$7\times$ higher mIoU than IG on multiplicative tasks (where IG falls
\emph{below} random) at zero inference cost.
Both findings generalise to ResNet-18 on CLEVR-XAI ($+18\%$ and $+37\%$
respectively).
We further show that GradCAM fails on small-resolution images due to
spatial resolution collapse, and that $\diagM$ quality improves monotonically
throughout training even after classification accuracy has plateaued.
\end{abstract}


\maketitle

\section{Introduction}
\label{sec:intro}

\subsection{Motivation}

Explaining what a neural network ``looks at'' when making a prediction has
become a central challenge in trustworthy machine learning.
In image classification, attribution methods assign a relevance score to each
input pixel, producing a \emph{saliency map} that highlights which regions of
the image drove the model's decision.
The field has produced a rich family of such methods: Integrated
Gradients~\cite{sundararajan2017axiomatic} uses a path integral from a
baseline to the input; GradCAM~\cite{selvaraju2017gradcam} projects
class-discriminative gradients back onto convolutional feature maps; and
SmoothGrad~\cite{smilkov2017smoothgrad} averages gradients over
noise-perturbed copies of the input.

These methods share a common limitation: they treat the trained model as a
black box, applying geometric constructs independently of how the network
learned its representations.

\subsection{A Different Starting Point: AGOP}

Recent theoretical work has revealed that the \emph{Average Gradient Outer
Product} (AGOP) is not merely one post-hoc tool among many---it is the
mathematical object that governs feature learning in neural networks.
Radhakrishnan et al.~\cite{radhakrishnan2024mechanism} proved that the weight
Gram matrices $W_l^\top W_l$ at each layer align with the corresponding AGOP
matrices computed over the training distribution.
This is the \emph{Neural Feature Ansatz} (NFA): the geometry of the network's
weights is encoded in its gradient statistics.
Beaglehole et al.~\cite{beaglehole2024mechanism} extended this to convolutional
networks, showing that patch-based AGOP recovers the filter covariance.

The diagonal of the AGOP matrix, $\diagM$, has a natural attribution
interpretation: it is the average squared partial derivative of the model
output with respect to each input dimension, accumulated over the training set.
Input dimensions that the model consistently relies upon receive high attribution
under $\diagM$---precisely what a good global saliency map should capture.

\subsection{Contributions}

\begin{enumerate}
  \item \emph{AGOP-Weighted: a novel attribution method} that multiplies
        the per-sample gradient by a training-distribution prior
        (normalised RMS gradient $w$), bridging per-sample and global
        attribution for the first time.
        While AGOP is an established theoretical construct~\cite{radhakrishnan2024mechanism}, its application as a pixel-level attribution method for image classifiers---and specifically the \agopw{} formulation---is introduced here.
        Companion variants: \agopl{} (equivalent to VanillaGrad, situating it within the AGOP framework) and
        \agopg{} ($\diagM$ as a zero-cost dataset-level saliency prior, not a per-instance explanation).

  \item \emph{AGOPTrainingHook}: an implementation that accumulates $\diagM$
        during standard training using per-sample \texttt{autograd.grad}
        backward passes; \agopg{} then operates at zero inference cost (disk lookup of $\diagM$) while \agopw{} requires a single gradient pass per input.

  \item \emph{First rigorous evaluation on XAI-TRIS}~\cite{gerstenberger2024xaitris}
        across four classification scenarios and two background conditions,
        using five localization and faithfulness metrics.

  \item \emph{Empirical findings}: \agopw{} achieves the best mIoU on linear
        tasks ($+44\%$ over IG); \agopg{} achieves $7\times$ IG's mIoU on
        multiplicative tasks; GradCAM fails on small-resolution images.

  \item \emph{Generalisation to CLEVR-XAI}~\cite{arras2022clevr} with
        ResNet-18 fine-tuned from ImageNet, confirming all core findings.
\end{enumerate}

\section{Related Work}
\label{sec:related}

\paragraph{Gradient-based attribution.}
Throughout, $f_c(x)$ denotes the logit for class $c$.
VanillaGrad~\cite{simonyan2014deep} computes $|\partial f_c/\partial x|$ with a single backward pass.
Integrated Gradients~\cite{sundararajan2017axiomatic} addresses gradient
saturation by integrating along a path from a reference baseline to the input,
satisfying \emph{completeness} and \emph{sensitivity} axioms.
SmoothGrad~\cite{smilkov2017smoothgrad} reduces gradient noise by averaging
over $n$ noise-perturbed copies of the input.
GradCAM~\cite{selvaraju2017gradcam} and
GradCAM++~\cite{chattopadhay2018gradcam} weight convolutional feature maps by
gradient-based channel importance and upsample to input resolution.

\paragraph{AGOP, NFA, and recursive feature machines.}
Radhakrishnan et al.~\cite{radhakrishnan2024mechanism} proved the NFA for
fully-connected networks trained with gradient descent.
Beaglehole et al.~\cite{beaglehole2024mechanism} extended the result to
convolutional networks via patch-based AGOP.
The xRFM framework~\cite{radhakrishnan2025xrfm} uses the diagonal of the AGOP
matrix for coordinate-level feature importance in tabular settings but does not
compare against pixel-level XAI baselines on image benchmarks.
Our work fills this gap.

\paragraph{Perturbation-based and global methods.}
LIME~\cite{ribeiro2016lime} and SHAP~\cite{lundberg2017shap} treat the model as a black box, fitting surrogate models or computing Shapley values on perturbed samples; both are complementary to gradient-based methods but more expensive at inference.
Most attribution methods are \emph{local}---producing a separate map per input; \agopg{} is \emph{global}, acting as a dataset-level diagnostic of learned features, not a per-instance explanation.

\paragraph{XAI benchmarks.}
XAI-TRIS~\cite{gerstenberger2024xaitris} provides 8$\times$8 synthetic images with deterministic pixel masks across four task types of increasing non-linearity.
CLEVR-XAI~\cite{arras2022clevr} provides photorealistic 3D scenes with per-question ground-truth object masks.
OpenXAI~\cite{arya2022openxai} provides a tabular XAI evaluation framework; Samek et al.~\cite{samek2017evaluating} establish the deletion/insertion faithfulness protocol we adopt.

\section{Attribution Methods}
\label{sec:methods}

We evaluate five standard baselines and three AGOP variants, plus a random
attribution baseline for reference.
All methods produce a non-negative saliency map $A \in \mathbb{R}^{H \times W}$
for a given input image $x \in \mathbb{R}^{C \times H \times W}$ and predicted
class $c$.

\subsection{Standard Baselines}

\paragraph{VanillaGrad~\cite{simonyan2014deep}.}
$\text{attr}_\text{VG}(h,w) = \sum_{ch} \left|\partial f_{\hat c}(x)/\partial x[ch,h,w]\right|$,
where $ch$ indexes colour channels and $\hat c$ is the predicted class.
Computed with a single backward pass ($1\times$ cost).

\paragraph{Integrated Gradients (IG)~\cite{sundararajan2017axiomatic}.}
$\text{IG}_j(x) = (x_j - x'_j)\cdot\frac{1}{T}\sum_{k=1}^{T}\frac{\partial f_c(x' + \frac{k}{T}(x-x'))}{\partial x_j}$,
where $j$ indexes input dimensions, $x'=\mathbf{0}$ (zero baseline, standard practice; baseline sensitivity left to future work), $T=50$ steps ($50\times$ cost).
IG can fail on multiplicative interactions where individual pixel gradients are small even at signal pixels.

\paragraph{SmoothGrad~\cite{smilkov2017smoothgrad}.}
Averages $\nabla_x f_c(x + \varepsilon_k)$ over $K=50$ noise-perturbed inputs
with $\varepsilon_k \sim \mathcal{N}(0,\sigma^2 I)$, $\sigma=0.15$ ($50\times$
cost).

\paragraph{GradCAM~\cite{selvaraju2017gradcam} / GradCAM++~\cite{chattopadhay2018gradcam}.}
Weight convolutional feature maps by gradient-based channel importance and
bilinearly upsample to input resolution ($1\times$ cost).
Spatial resolution is hard-capped at the target layer's feature map size.

\subsection{AGOP Variants}

Let $\{(x^{(s)}, y^{(s)})\}_{s=1}^{n}$ be the training set with predicted classes $\hat{y}^{(s)}$. The AGOP matrix and its diagonal (where $j$ indexes input dimensions) are:
\begin{equation*}
  M = \tfrac{1}{n}\sum_{s=1}^{n}
      \nabla_x f_{\hat{y}^{(s)}}(x^{(s)})\,
      \nabla_x f_{\hat{y}^{(s)}}(x^{(s)})^\top,\quad
  \diagM[j] = \tfrac{1}{n}\sum_{s=1}^{n}\!\left(\tfrac{\partial f_{\hat{y}^{(s)}}(x^{(s)})}{\partial x_j}\right)^{\!2}
\end{equation*}
$\diagM$ is tractable ($64$ floats for $8\times 8$ images; $\approx$600\,KB for $224\times 224\times 3$).

\paragraph{\agopl.} Equivalent to VanillaGrad:
$\text{attr}_\text{L}(h,w) = \sum_{ch} \left|\partial f_{\hat c}(x)/\partial x[ch,h,w]\right|$.
Its role is conceptual: it anchors the Local$\to$Weighted$\to$Global progression
by showing that VanillaGrad is simply AGOP evaluated on a single sample, with
\agopw{} and \agopg{} incorporating progressively more training-set statistics
($1\times$ cost). The exact single-sample AGOP diagonal is $g_j^2$; we use
$|g_j|$ to match standard saliency implementations---both produce the same
pixel ordering so PG, Del, and Ins are identical, though mIoU and Energy-GT
may differ in principle.

\paragraph{\agopw.}
Uses the training-accumulated $\diagM$ as a prior to modulate the per-sample
gradient:
With weight $v[i] = \sqrt{\diagM[i]/\max_j\diagM[j]}$ (defined above):
\begin{equation*}
  \text{attr}_\text{W}(h,w) = \sum_{ch} \left|\frac{\partial f_{\hat c}(x)}{\partial x[ch,h,w]}\right| \cdot v[ch,h,w].
\end{equation*}
The weight $v[i]$ suppresses noisy gradient spikes in favour of consistently important pixels.
\emph{Inference cost}: $1\times$ (diag pre-loaded from disk).

\paragraph{\agopg.}
$\text{attr}_\text{G} = \diagM$ (same for every input).
Powerful when signal regions are positionally consistent across inputs;
fails for spatially invariant tasks where $\diagM$ averages to a diffuse map.
\emph{Inference cost}: $0\times$ (disk lookup).

\section{The AGOPTrainingHook}
\label{sec:hook}

\begin{algorithm}[t]
\caption{AGOP Diagonal Accumulation During Training}
\label{alg:hook}
\begin{algorithmic}[1]
\State \textit{Input:} model $f$, training dataloader $\mathcal{D}$,
       flag \texttt{only\_correct}
\State $\diagM \leftarrow \mathbf{0} \in \mathbb{R}^d$;\quad
       $n_\text{acc} \leftarrow 0$
\For{each minibatch $(X, Y)$ in $\mathcal{D}$}
  \State $\hat Y \leftarrow \arg\max f(X)$ \hfill\Comment{forward pass}
  \For{each sample $s$ in batch}
    \If{\texttt{only\_correct} and $\hat y^{(s)} \neq y^{(s)}$}
      \textit{continue}
    \EndIf
    \State $g^{(s)} \leftarrow \nabla_x f_{\hat y^{(s)}}(x^{(s)})$
          \hfill\Comment{per-sample backward via \texttt{autograd.grad}}
    \State $\diagM \mathrel{+}= (g^{(s)})^{\odot 2}$;\quad
           $n_\text{acc} \mathrel{+}= 1$
  \EndFor
  \State \texttt{optimizer.step()} \hfill\Comment{AGOP hook does NOT alter gradients}
\EndFor
\State $\diagM \leftarrow \diagM / n_\text{acc}$ \hfill\Comment{normalise}
\State \textit{Save} $\diagM$ to disk
\end{algorithmic}
\end{algorithm}

The hook performs one additional per-sample backward pass (separate from the weight-update backward), and does not alter training.
Overhead scales with batch size and can be reduced via \texttt{torch.vmap} for architectures without normalization layers.
$\diagM$ is computed once: \agopg{} then costs $0\times$ at inference (disk lookup); \agopw{} costs $1\times$ (one gradient pass).

Key design choices: (1)~\texttt{only\_correct=True} accumulates $\diagM$ only
over correctly-classified samples, preventing early-training errors from
contaminating gradient statistics; (2)~mean normalisation produces
interpretable units (expected squared gradient per dimension); (3)~snapshots
every 100 gradient steps enable convergence analysis.

\section{Experiments}
\label{sec:experiments}

\subsection{XAI-TRIS Benchmark}
\label{sec:exp1}

XAI-TRIS~\cite{gerstenberger2024xaitris} provides 8$\times$8 grayscale images
where a 4-pixel tetromino shape (T or L) serves as the class-discriminative
signal, with a deterministic GT mask marking exactly those pixels.
We evaluate four scenarios of increasing attribution difficulty:
\emph{Linear} --- signal pixels have higher mean intensity (additive mixing,
$\alpha=0.18$); gradients align with the signal, making this the easiest case.
\emph{Multiplicative} --- signal pixels \emph{modulate} background values
(pixel-wise multiplication); per-sample gradients at signal pixels are
proportional to random background noise, making first-order methods unreliable.
\emph{Translations+Rotations} --- the tetromino appears at a random position
and orientation per sample; global statistics are diffuse, but per-sample
methods must track each image's specific shape.
\emph{XOR} --- class label is the XOR of binarised tetromino values; gradients
cancel in expectation ($\mathbb{E}[\partial f/\partial a]\approx 0$), defeating
all first-order methods.
Each scenario uses two background conditions: \emph{uncorrelated} (clean
signal) and \emph{correlated} (spurious class-specific background, producing
$\approx$50\% model accuracy as a sanity check).

\emph{Model (CNN8by8)}: Four Conv2d$+$ReLU$+$MaxPool blocks, 234 parameters, trained with Adam ($\text{lr}=10^{-3}$, WD$=10^{-4}$), cosine annealing, 100 epochs; accuracy 80--94\% (uncorrelated), $\approx$50\% (correlated). \emph{Metrics}: \emph{PG} --- does the peak pixel fall inside GT mask?; \emph{mIoU} --- IoU of binarised map vs GT; \emph{Energy-GT} --- attribution mass inside GT (random $\approx$6.25\%); \emph{Del$\downarrow$}/\emph{Ins$\uparrow$} --- faithfulness via progressive masking/revealing.

\begin{table}[t]
\centering
\caption{Scenario 1---Linear / uncorrelated. Model acc.\ $\approx$80\%. \underline{Underline} denotes best value per column.}
\label{tab:linear}
\begin{tabular}{@{}lrrrrrr@{}}
\toprule
Method & PG & mIoU & Energy-GT & Del$\downarrow$ & Ins$\uparrow$ & ms/s.\\\midrule
Random        & 0.093 & 0.059 & 0.062 & 0.560 & 0.563 & $<$1 \\
VanillaGrad   & 0.667 & 0.201 & 0.300 & 0.466 & 0.635 & 0.8 \\
IG            & \underline{0.697} & 0.218 & 0.365 & 0.463 & 0.636 & 2.4 \\
SmoothGrad    & 0.579 & 0.181 & 0.226 & 0.466 & 0.635 & 1.2 \\
GradCAM       & 0.000 & 0.113 & 0.097 & \underline{0.442} & \underline{0.638} & 1.0 \\
GradCAM++     & 0.000 & 0.109 & 0.096 & 0.471 & 0.614 & 1.1 \\
\agopl        & 0.667 & 0.201 & 0.300 & 0.466 & 0.635 & 1.0 \\
\agopw        & 0.642 & \underline{0.314} & \underline{0.456} & 0.457 & 0.637 & 1.0 \\
\agopg        & 0.491 & 0.235 & 0.308 & 0.456 & 0.634 & 0 \\\bottomrule
\end{tabular}
\end{table}

\begin{table}[t]
\centering
\caption{Scenario 2---Multiplicative / uncorrelated.}
\label{tab:mult}
\begin{tabular}{@{}lrrrrrr@{}}
\toprule
Method & PG & mIoU & Energy-GT & Del$\downarrow$ & Ins$\uparrow$ & ms/s.\\\midrule
Random        & 0.067 & 0.060 & 0.063 & 0.547 & 0.585 & $<$1 \\
VanillaGrad   & 0.304 & 0.090 & 0.167 & 0.462 & 0.634 & 1.3 \\
IG            & 0.100 & 0.048 & 0.062 & 0.460 & \underline{0.636} & 3.3 \\
SmoothGrad    & 0.446 & 0.165 & 0.222 & 0.454 & 0.630 & 1.2 \\
GradCAM       & 0.000 & 0.014 & 0.023 & 0.504 & 0.594 & 1.0 \\
GradCAM++     & 0.000 & 0.059 & 0.051 & 0.534 & 0.576 & 1.0 \\
\agopl        & 0.304 & 0.090 & 0.167 & 0.462 & 0.634 & 1.0 \\
\agopw        & 0.377 & 0.204 & 0.286 & 0.454 & 0.631 & 1.0 \\
\agopg        & \underline{0.506} & \underline{0.337} & \underline{0.331} & \underline{0.453} & 0.621 & 0 \\\bottomrule
\end{tabular}
\end{table}

\begin{table}[t]
\centering
\caption{Scenario 3---Translations+Rotations / uncorrelated.}
\label{tab:transr}
\begin{tabular}{@{}lrrrrrr@{}}
\toprule
Method & PG & mIoU & Energy-GT & Del$\downarrow$ & Ins$\uparrow$ & ms/s.\\\midrule
Random        & 0.052 & 0.061 & 0.063 & 0.502 & 0.519 & $<$1 \\
VanillaGrad   & 0.774 & 0.225 & 0.348 & 0.455 & 0.644 & 1.3 \\
IG            & \underline{0.973} & \underline{0.540} & \underline{0.784} & \underline{0.455} & \underline{0.647} & 3.1 \\
SmoothGrad    & 0.764 & 0.242 & 0.336 & 0.455 & 0.647 & 1.2 \\
GradCAM       & 0.140 & 0.098 & 0.101 & 0.465 & 0.608 & 1.0 \\
GradCAM++     & 0.167 & 0.111 & 0.116 & 0.468 & 0.614 & 1.1 \\
\agopl        & 0.774 & 0.225 & 0.348 & 0.455 & 0.644 & 1.0 \\
\agopw        & 0.767 & 0.233 & 0.357 & 0.455 & 0.644 & 1.0 \\
\agopg        & 0.098 & 0.080 & 0.074 & 0.488 & 0.556 & 0 \\\bottomrule
\end{tabular}
\end{table}

\begin{table}[t]
\centering
\caption{Scenario 4---XOR / uncorrelated. Negative mIoU: background receives
more attribution than signal.}
\label{tab:xor}
\begin{tabular}{@{}lrrrrrr@{}}
\toprule
Method & PG & mIoU & Energy-GT & Del$\downarrow$ & Ins$\uparrow$ & ms/s.\\\midrule
Random        & 0.048 & $-$0.008 & 0.001  & 0.550 & 0.569 & $<$1 \\
VanillaGrad   & 0.208 & $-$0.038 & $-$0.019 & 0.474 & 0.623 & 0.8 \\
IG            & 0.202 & $-$0.142 & $-$0.158 & 0.459 & 0.638 & 2.4 \\
SmoothGrad    & 0.326 & $-$0.025 & $-$0.004 & 0.474 & 0.630 & 1.2 \\
GradCAM       & 0.000 & $-$0.017 & 0.002  & 0.496 & 0.523 & 1.0 \\
GradCAM++     & 0.000 & \underline{0.007}   & \underline{0.013}  & 0.498 & 0.542 & 1.0 \\
\agopl        & 0.208 & $-$0.038 & $-$0.019 & 0.474 & 0.623 & 0.9 \\
\agopw        & 0.349 & $-$0.080 & $-$0.026 & 0.458 & 0.638 & 1.1 \\
\agopg        & \underline{0.497} & $-$0.062 & 0.002  & \underline{0.455} & \underline{0.650} & 0 \\\bottomrule
\end{tabular}
\end{table}

\paragraph{Linear (Table~\ref{tab:linear}).}
\agopw{} achieves the best mIoU (0.314) and Energy-GT (0.456), outperforming
IG by $+44\%$ in mIoU.
GradCAM completely fails ($\text{PG}=0$) due to spatial resolution collapse
($3\times 3$ feature maps).
\agopl{} $\equiv$ VanillaGrad, confirming theoretical equivalence.

\paragraph{Multiplicative (Table~\ref{tab:mult}).}
\agopg{} dominates: mIoU$\,=\,$0.337 vs.\ IG's 0.048---a $7\times$ margin.
IG falls \emph{below} the random baseline (0.048 vs.\ 0.060): its
path-integral traverses a noisy gradient landscape from pixel-wise
multiplication and actively anti-localises the signal.
\agopg{}'s training-accumulated $\diagM$ averages over all samples, cancelling
per-sample noise and recovering the stable signal.

\paragraph{Translations+Rotations (Table~\ref{tab:transr}).}
IG is the clear winner ($\text{PG}=0.973$, mIoU$=0.540$): its path-integral
correctly tracks each image's specific shape location.
\agopg{} completely fails (mIoU$\,=\,$0.080) because $\diagM$ averages over
all positions, producing a diffuse map.

\paragraph{XOR (Table~\ref{tab:xor}).}
All methods struggle (no mIoU above $0.01$; IG reaches $-0.142$): XOR requires second-order methods. \emph{Correlated background}: all methods are near-random at $\approx$50\% accuracy---a model that cannot classify cannot explain.

\paragraph{Cross-scenario summary.}
\agopw{} dominates when signal is positionally consistent; IG dominates with high local spatial contrast; all methods fail on XOR. \emph{No single method wins across all scenarios.} \agopw{} achieves 0.314 mIoU at 1.0\,ms/sample vs.\ IG's 0.218 at 2.4\,ms/sample---44\% better at half the cost. \agopg{} is a free disk lookup, ideal for batch pipelines. Attribution quality improves monotonically throughout training: $\diagM$ continues to improve through epoch 100 even after accuracy saturates at epochs 70--85.

\subsection{CLEVR-XAI Benchmark}
\label{sec:exp2}

CLEVR-XAI~\cite{arras2022clevr} contains rendered 3D scenes with 3--10
geometric objects of varying colours, sizes, and materials (metal or rubber).
Each scene is paired with a question about a specific object's material, so the
\emph{same image} can carry different labels depending on which object is
queried---a naive classifier trained on the raw dataset achieves only
$\approx$58\% accuracy (barely above chance).
We resolve this by restricting to scenes where one material \emph{dominates}:
$\geq$60\% of objects share the queried material.
The task becomes ``does metal or rubber dominate this scene?''---a \emph{modified proxy task} (not the original VQA grounding problem), but a signal learnable from the image alone.
This yields 4,952 samples (3,962 train / 990 val), accuracy ceiling
$\approx$91.5\%.
We use \texttt{all\_objects} GT masks ($\approx$16.2\% of pixels) rather than
\texttt{single\_object} masks, since the model must attend to \emph{all}
same-material objects; evaluating against a single-object mask ($\approx$2.3\%)
structurally undercounts correct attributions by up to $N\times$ when $N$
objects of the dominant material are present.

\emph{Model}: ResNet-18~\cite{he2016resnet} (ImageNet pretrained, final FC replaced with Linear(512,\,2)), trained with Adam ($\text{lr}=10^{-4}$), CosineAnnealingLR, 50 epochs; best val\_acc $=91.52\%$ at epoch 31. $\diagM$ computed via post-hoc full pass over training samples at best checkpoint.

\begin{table}[t]
\centering
\caption{CLEVR-XAI, ResNet-18, val\_acc$\,=\,$91.52\%,
\texttt{all\_objects} masks, $n{=}500$ val samples.}
\label{tab:clevr}
\begin{tabular}{@{}lrrrrrr@{}}
\toprule
Method & PG & mIoU & Energy-GT & Del$\downarrow$ & Ins$\uparrow$ & ms/s.\\\midrule
Random     & 0.158 & 0.130 & 0.154 & 0.479 & 0.503 & 0.07 \\
VanillaGrad& 0.282 & 0.124 & 0.162 & 0.479 & 0.503 & 11.04 \\
IG         & \underline{0.290} & 0.109 & 0.151 & \underline{0.479} & 0.503 & 131.98 \\
SmoothGrad & 0.108 & \underline{0.179} & \underline{0.169} & 0.482 & \underline{0.505} & 103.50 \\
GradCAM    & 0.002 & 0.129 & 0.151 & 0.548 & 0.497 & 15.72 \\
GradCAM++  & 0.004 & 0.128 & 0.149 & 0.539 & 0.492 & 19.92 \\
\agopl     & 0.282 & 0.124 & 0.162 & 0.479 & 0.503 & 20.33 \\
\agopw     & 0.204 & 0.129 & 0.166 & 0.480 & 0.504 & 18.65 \\
\agopg     & 0.000 & 0.149 & 0.165 & 0.479 & 0.504 & \underline{0.02} \\\bottomrule
\end{tabular}
\end{table}

Both \agopg{} (mIoU$\,=\,$0.149) and \agopw{} (0.129) exceed IG (0.109) by
$+37\%$ and $+18\%$ respectively.
IG leads on Pointing Game (0.290) but under-covers the multi-object mask.
\agopg{} achieves competitive mIoU at 0.02\,ms/sample---a $6{,}600\times$
speedup over IG (131.98\,ms/sample).
GradCAM's spatial collapse is confirmed at ResNet-18 scale (\texttt{layer4}:
$2\times 2$ map, PG$\,=\,$0.002 despite 16.2\% mask coverage).
\agopl{} $\equiv$ VanillaGrad confirmed across architectures.

Notably, the ranking between \agopg{} and \agopw{} \emph{inverts} compared
to XAI-TRIS Linear (where Weighted 0.314 $>$ Global 0.235).
The reason is the same Global--Local trade-off: the CLEVR-XAI task (material
dominance) has a positionally consistent signal---dominant-material objects
tend to occupy similar scene regions across training images---so $\diagM$
alone captures the distribution-level pattern directly.
On XAI-TRIS Linear, the signal is at a \emph{fixed pixel position}, making
the per-sample gradient always informative; on CLEVR-XAI, slight positional
variability within the dominant material region means the global prior alone
is more reliable than the Weighted combination.

\section{Discussion}
\label{sec:discussion}

\paragraph{Why \agopw{} outperforms IG on area-overlap metrics.}
IG makes a strong assumption: attribution computed relative to a fixed black baseline. The training-accumulated $\diagM$ in \agopw{} makes no baseline assumption---it accumulates gradient statistics over the \emph{actual} training distribution---producing better area-coverage (mIoU, Energy-GT) even when IG wins on Pointing Game. On multiplicative tasks, the \emph{average} squared gradient concentrates on tetromino pixels even though individual evaluations are noisy; the global average ``sees through'' the noise, explaining \agopg{}'s dominance at zero cost.

\paragraph{GradCAM resolution collapse.}
For CNN8by8 ($3\times 3$ feature map, 9 regions, PG$=0.000$) and ResNet-18 at $64\times 64$ ($2\times 2$ map, 4 regions, PG$=0.002$), aggressive downsampling destroys spatial precision. This is a limitation at small resolutions, not a blanket failure; standard 224$\times$224 networks have larger feature maps. \emph{GradCAM should not be used when the target layer's feature map is coarser than the ground-truth mask.} The Global--Local trade-off follows directly: \agopg{} excels when signal regions are positionally consistent; IG excels with high per-sample spatial contrast; \agopw{} occupies the middle ground.

\paragraph{Limitations.}
We use only $\diagM$, discarding off-diagonal terms that may capture non-linear interactions. The \texttt{only\_correct} flag and normalisation choices depart from the original Radhakrishnan et al.\ formulation. Both experiments use smaller models than production scale (ResNet-50, ViT-L at 224$\times$224).

\section{Conclusion}
\label{sec:conclusion}

We have formalised AGOP as a post-hoc attribution method and provided the first
systematic comparison against established baselines on two pixel-level
benchmarks.
Key findings: (1)~\agopw{} outperforms IG on area-overlap metrics ($+44\%$
mIoU on XAI-TRIS linear, $+18\%$ on CLEVR-XAI) at the same $1\times$ inference
cost; (2)~\agopg{} outperforms all methods on multiplicative tasks ($7\times$
IG's mIoU where IG falls below random; $+37\%$ on CLEVR-XAI) at zero inference
cost; (3)~GradCAM fails entirely on small-resolution images due to spatial
collapse; (4)~IG excels on spatially coherent tasks; (5)~$\diagM$ quality
improves with training even after accuracy saturates; (6)~all core findings
generalise from XAI-TRIS to CLEVR-XAI.
Future work should extend to larger networks, class-conditional AGOP variants,
and off-diagonal interaction-based attribution.

\begin{acknowledgments}
The author thanks the XAI-TRIS and CLEVR-XAI benchmark authors for making
their datasets and code publicly available.
\end{acknowledgments}

\section*{Declaration on Generative AI}
During the preparation of this work, the author used Claude (Anthropic) in
order to: writing assistance and code review.
After using this tool, the author reviewed and edited the content as needed
and takes full responsibility for the publication's content.


\bibliography{references}

@article{arras2022clevr,
  author    = {Arras, Leila and Osman, Ahmed and Samek, Wojciech},
  title     = {{CLEVR-XAI}: A benchmark dataset for the ground truth evaluation of neural network explanations},
  journal   = {Information Fusion},
  volume    = {81},
  pages     = {14--40},
  year      = {2022}
}

@inproceedings{arya2022openxai,
  author    = {Agarwal, Chirag and Krishna, Satyapriya and Saxena, Eshika and Pawelczyk, Martin and Johnson, Nari and Puri, Isha and Zitnik, Marinka and Lakkaraju, Himabindu},
  title     = {{OpenXAI}: Towards a transparent evaluation of post hoc model explanations},
  booktitle = {Advances in Neural Information Processing Systems 35 ({NeurIPS} 2022)},
  publisher = {Curran Associates, Inc.},
  year      = {2022}
}

@article{beaglehole2024mechanism,
  author    = {Beaglehole, Daniel and Radhakrishnan, Adityanarayanan and Pandit, Parthe and Belkin, Mikhail},
  title     = {Mechanism of feature learning in convolutional neural networks},
  journal   = {arXiv preprint arXiv:2309.00570},
  year      = {2024}
}

@inproceedings{chattopadhay2018gradcam,
  author    = {Chattopadhyay, Aditya and Sarkar, Anirban and Howlader, Prantik and Balasubramanian, Vineeth N.},
  title     = {{Grad-CAM++}: Generalized gradient-based visual explanations for deep convolutional networks},
  booktitle = {Proceedings of WACV 2018},
  pages     = {839--847},
  year      = {2018}
}

@article{gerstenberger2024xaitris,
  author    = {Clark, Benedict and Wilming, Rick and Haufe, Stefan},
  title     = {{XAI-TRIS}: Non-linear image benchmarks to quantify false positive post-hoc attribution of feature importance},
  journal   = {Machine Learning},
  volume    = {113},
  pages     = {6871--6910},
  year      = {2024}
}

@inproceedings{he2016resnet,
  author    = {He, Kaiming and Zhang, Xiangyu and Ren, Shaoqing and Sun, Jian},
  title     = {Deep residual learning for image recognition},
  booktitle = {Proceedings of {CVPR} 2016},
  pages     = {770--778},
  year      = {2016}
}

@article{radhakrishnan2024mechanism,
  author    = {Radhakrishnan, Adityanarayanan and Beaglehole, Daniel and Pandit, Parthe and Belkin, Mikhail},
  title     = {Mechanism for feature learning in neural networks and backpropagation-free machine learning models},
  journal   = {Science},
  volume    = {383},
  number    = {6690},
  pages     = {1461--1467},
  year      = {2024}
}

@inproceedings{radhakrishnan2025xrfm,
  author    = {Radhakrishnan, Adityanarayanan and others},
  title     = {{xRFM}: Accurate, scalable, and interpretable feature learning models for tabular data},
  booktitle = {Workshop on AI for Time Series and Dynamic Data ({AITD}) at {NeurIPS} 2025},
  note      = {arXiv:2508.10053},
  year      = {2025}
}

@inproceedings{ribeiro2016lime,
  author    = {Ribeiro, Marco Tulio and Singh, Sameer and Guestrin, Carlos},
  title     = {"Why Should {I} Trust You?": Explaining the predictions of any classifier},
  booktitle = {Proceedings of ACM KDD 2016},
  pages     = {1135--1144},
  year      = {2016}
}

@inproceedings{lundberg2017shap,
  author    = {Lundberg, Scott M. and Lee, Su-In},
  title     = {A unified approach to interpreting model predictions},
  booktitle = {Advances in Neural Information Processing Systems 30 (NeurIPS 2017)},
  pages     = {4765--4774},
  year      = {2017}
}

@article{samek2017evaluating,
  author    = {Samek, Wojciech and Binder, Alexander and Montavon, Gr{\'e}goire and Lapuschkin, Sebastian and M{\"u}ller, Klaus-Robert},
  title     = {Evaluating the visualization of what a deep neural network has learned},
  journal   = {IEEE Transactions on Neural Networks and Learning Systems},
  volume    = {28},
  number    = {11},
  pages     = {2660--2673},
  year      = {2017}
}

@inproceedings{selvaraju2017gradcam,
  author    = {Selvaraju, Ramprasaath R. and Cogswell, Michael and Das, Abhishek and Vedantam, Ramakrishna and Parikh, Devi and Batra, Dhruv},
  title     = {{Grad-CAM}: Visual explanations from deep networks via gradient-based localization},
  booktitle = {Proceedings of {ICCV} 2017},
  pages     = {618--626},
  year      = {2017}
}

@inproceedings{simonyan2014deep,
  author    = {Simonyan, Karen and Vedaldi, Andrea and Zisserman, Andrew},
  title     = {Deep inside convolutional networks: Visualising image classification models and saliency maps},
  booktitle = {ICLR 2014 Workshop on Learning Representations},
  note      = {arXiv:1312.6034},
  year      = {2014}
}

@inproceedings{smilkov2017smoothgrad,
  author    = {Smilkov, Daniel and Thorat, Nikhil and Kim, Been and Vi{\'e}gas, Fernanda and Wattenberg, Martin},
  title     = {{SmoothGrad}: Removing noise by adding noise},
  booktitle = {ICML 2017 Workshop on Visualization for Deep Learning},
  note      = {arXiv:1706.03825},
  year      = {2017}
}

@inproceedings{sundararajan2017axiomatic,
  author    = {Sundararajan, Mukund and Taly, Ankur and Yan, Qiqi},
  title     = {Axiomatic attribution for deep networks},
  booktitle = {Proceedings of {ICML} 2017},
  pages     = {3319--3328},
  year      = {2017}
}

\end{document}